\documentclass[10pt,twocolumn,letterpaper]{article}

\usepackage{cvpr}
\usepackage{times}
\usepackage{epsfig}
\usepackage{graphicx}
\usepackage{amsmath}
\usepackage{amssymb}

\usepackage{color}
\usepackage{multirow}
\usepackage{booktabs}
\usepackage{enumerate}
\usepackage{cite}
\usepackage{algorithm}
\usepackage{algpseudocode}
\usepackage{bbm}
\usepackage{subfigure}
\usepackage{boldline}
\usepackage{enumitem}
\usepackage{helvet} 
\usepackage{courier} 
\usepackage[hyphens]{url} 

\usepackage{paralist}

\usepackage[pagebackref=true,breaklinks=true,letterpaper=true,colorlinks,bookmarks=false]{hyperref}

\cvprfinalcopy 

\def\cvprPaperID{2062} 

\ifcvprfinal\fi
\begin{document}

\title{Transferable, Controllable, and Inconspicuous Adversarial Attacks on Person Re-identification With Deep Mis-Ranking}

\newcommand*{\affaddr}[1]{#1} 
\newcommand*{\affmark}[1][*]{\textsuperscript{#1}}
\newcommand*{\email}[1]{\texttt{#1}}

\author{%
Hongjun Wang\affmark[1]\affmark[\textasteriskcentered]\qquad Guangrun Wang\affmark[1]\thanks{Equal contribution}\qquad Ya Li\affmark[2]\qquad Dongyu Zhang\affmark[2]\qquad Liang Lin\affmark[1,3]\thanks{Corresponding author} \\
\affaddr{\affmark[1]Sun Yat-sen University}\qquad
\affaddr{\affmark[2]Guangzhou University}\qquad
\affaddr{\affmark[3]DarkMatter AI}\\
\email{\tt\small \affmark[1]\{wanghq8,wanggrun,zhangdy27\}@mail2.sysu.edu.cn}\qquad
\email{\tt\small \affmark[2]liya@gzhu.edu.cn}\qquad
\email{\tt\small \affmark[3]linliang@ieee.org}
}

\maketitle

\begin{abstract}
The success of DNNs has driven the extensive applications of person re-identification (ReID) into a new era. However, whether ReID inherits the vulnerability of DNNs remains unexplored. To examine the robustness of ReID systems is rather important because the insecurity of ReID systems may cause severe losses, e.g., the criminals may use the adversarial perturbations to cheat the CCTV systems.

In this work, we examine the insecurity of current best-performing ReID models by proposing a learning-to-mis-rank formulation to perturb the ranking of the system output. As the cross-dataset transferability is crucial in the ReID domain, we also perform a back-box attack by developing a novel multi-stage network architecture that pyramids the features of different levels to extract general and transferable features for the adversarial perturbations. Our method can control the number of malicious pixels by using differentiable multi-shot sampling. To guarantee the inconspicuousness of the attack, we also propose a new perception loss to achieve better visual quality.

Extensive experiments on four of the largest ReID benchmarks (i.e., Market1501 \cite{zheng2015scalable}, CUHK03 \cite{li2014deepreid}, DukeMTMC \cite{ristani2016performance}, and MSMT17 \cite{wei2018person}) not only show the effectiveness of our method, but also provides directions of the future improvement in the robustness of ReID systems.
For example, the accuracy of one of the best-performing ReID systems drops sharply from 91.8\% to 1.4\% after being attacked by our method. Some attack results are shown in Fig. \ref{fig:intro_market}. The code is available at \url{https://github.com/whj363636/Adversarial-attack-on-Person-ReID-With-Deep-Mis-Ranking}.
\end{abstract}

\section{Introduction}\label{sect:intro}
The success of deep neural networks (DNNs) has benefited a wide range of computer vision tasks, such as person re-identification (ReID), a crucial task aiming at matching pedestrians across cameras. In particular, DNNs have benefited ReID in learning discriminative features and adaptive distance metrics for visual matching, which drives ReID to a new era \cite{DBLP:journals/corr/abs-1711-08184,sun2018beyond}. Thanks to DNNs, there have been extensive applications of ReID in video surveillance or criminal identification for public safety.

\begin{figure}[t]
\vspace{-11pt}
\centering
\includegraphics[width=0.45\textwidth]{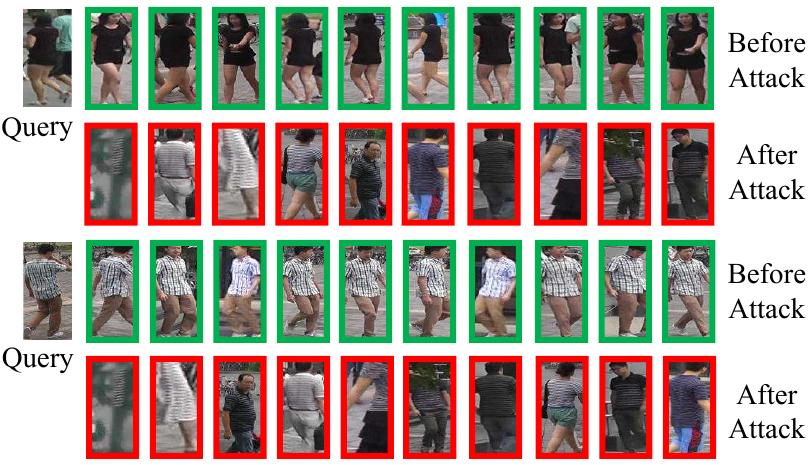}
\caption{\small{The rank-10 predictions of AlignedReID\cite{sun2018beyond} (one of the state-of-the-art ReID models) before and after our attack on Market-1501. The green boxes represent the correctly matching images, while the red boxes represent the mismatching images.}}\label{fig:intro_market}
\vspace{-18pt}
\end{figure}

Despite the impressive gain obtained from DNNs, whether ReID inherits the vulnerability of DNNs remains unexplored. Specifically, recent works found that DNNs are vulnerable to adversarial attacks \cite{DBLP:conf/iclr/LiuCLS17,su2019one} (An adversarial attack is to mislead a system with adversarial examples). In the past two years, the adversarial attack has achieved remarkable success in fooling DNN-based systems, e.g., image classification. Can the recent DNN-based ReID systems survive from an adversarial attack? The answer seems not promising. Empirically, evidence has shown that a person wearing bags, hats, or glasses can mislead a ReID system to output a wrong prediction\cite{xue2018clothing,DBLP:conf/bmvc/GouZRACS16,DBLP:conf/ijcnn/HuangWXZ19,li2014clothing,lin2019improving}. These examples may be regarded as natural adversarial examples.

To examine the robustness of ReID systems against adversarial attacks is of significant importance. Because the insecurity of ReID systems may cause severe losses, for example, in criminal tracking, the criminal may disguise themselves by placing adversarial perturbations (e.g., bags, hats, and glasses) on the most appropriate position of the body to cheat the video surveillance systems. By investigating the adversarial examples for the ReID systems, we can identify the vulnerability of these systems and help improve the robustness. For instance, we can identify which parts of a body are most vulnerable to the adversarial attack and require future ReID systems to pay attention to these parts. We can also improve ReID systems by using adversarial training in the future. In summary, developing adversarial attackers to attack ReID is desirable, although no work has been done before.


As the real-world person identities are endless, and the queried person usually does not belong to any category in the database, ReID is defined as a ranking problem rather than a classification problem. But existing attack methods for image classification, segmentation, detection, and face recognition do not fit a ranking problem. \textbf{Moreover}, since the image domains vary at different times and in different cameras, examining the robustness of ReID models by employing a cross-dataset black-box attack should also be taken into consideration. However, existing adversarial attack methods often have poor transferability, i.e., they are often designed for a sole domain of task (e.g., Dataset A) and can not be reused to another domain (e.g., Dataset B) due to their incapacity to find general representations for attacking. \textbf{Furthermore}, we focus on attacks that are inconspicuous to examine the insecurity of ReID models. Existing adversarial attack methods usually have a defective visual quality that can be perceived by humans.


To address the aforementioned issues, we design a transferable, controllable, and inconspicuous attacker to examine the insecurity of current best-performing ReID systems. We propose a learning-to-mis-rank formulation to perturb the ranking prediction of ReID models. A new mis-ranking loss function is designed to attack the ranking of the potential matches, which fits the ReID problem perfectly. Our mis-ranking based attacker is complementary to existing misclassification based attackers. Besides, as is suggested by \cite{Ilyas2019Adversarial_NIPS}, adversarial examples are features rather than bugs. Hence, to enhance the transferability of the attacker, one needs to improve the representation learning ability of the attacker to extract the general features for the adversarial perturbations. To this end, we develop a novel multi-stage network architecture for representation learning by pyramiding the features of different levels of the discriminator. This architecture shows impressive transferability in black-box attack for the complicated ReID tasks. The transferability leads to our joint solution of both white- and black-box attack. To make our attack inconspicuous, we improve the existing adversarial attackers in two aspects. \textbf{First}, the number of target pixels to be attacked is controllable in our method, due to the use of a differentiable multi-shot sampling. Generally, the adversarial attack can be considered as searching for a set of target pixels to be contaminated by noise. To make the search space continuous, we relax the choice of a pixel as a Gumbel softmax over all possible pixels. The number of target pixels is determined by the dynamic threshold of the softmax output and thus can be controllable. \textbf{Second}, a new perception loss is designed by us to improve the visual quality of the attacked images, which guarantees the inconspicuousness.

Experiments were performed on four of the largest ReID benchmarks, i.e., Market1501 \cite{zheng2015scalable}, CUHK03 \cite{li2014deepreid}, DukeMTMC \cite{ristani2016performance}, and MSMT17 \cite{wei2018person}. The results show the effectiveness of our method. For example, the performance of one of the best-performing systems \cite{DBLP:journals/corr/abs-1711-08184} drops sharply from 91.8\% to 1.4\% after attacked by our method. Except for showing a higher success attack rate, our method also provides interpretable attack analysis, which provides direction for improving the robustness and security of the ReID system. Some attack results are shown in Fig. \ref{fig:intro_market}. To summarize, our contribution is four-fold:

\vspace{3pt}
\begin{itemize}
\item{} To attack ReID, we propose a learning-to-mis-rank formulation to perturb the ranking of the system output. A new mis-ranking loss function is designed to attack the ranking of the predictions, which fits the ReID problem perfectly. Our mis-ranking based adversarial attacker is complementary to the existing misclassification based attackers.
\item{} To enhance the transferability of our attacker and perform a black-box attack, we improve the representation capacity of the attacker to extract general and transferable features for the adversarial perturbations.
\item{} To guarantee the inconspicuousness of the attack, we propose a differentiable multi-shot sampling to control the number of malicious pixels and a new perception loss to achieve better visual quality.
\item{} By using the above techniques, we examine the insecurity of existing ReID systems against adversarial attacks. Experimental validations on four of the largest ReID benchmarks show not only the successful attack and the visual quality but also the interpretability of our attack, which provides directions for the future improvement in the robustness of ReID systems.
\end{itemize}

\begin{figure*}[t]
\centering
\includegraphics[width=1.0\textwidth]{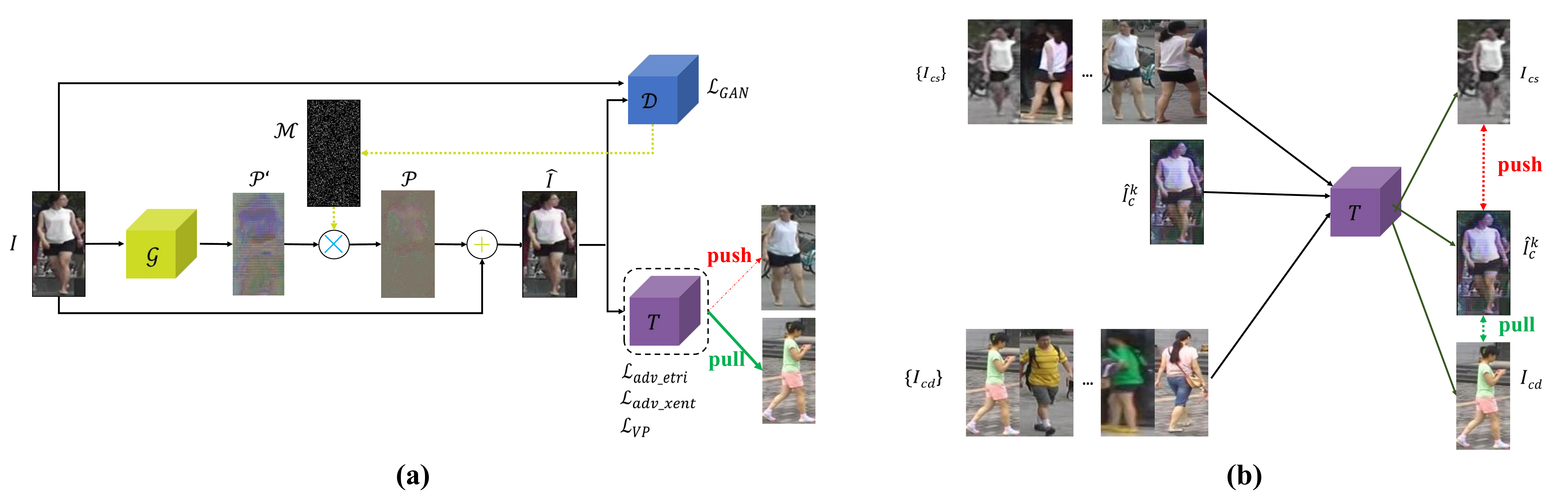}
\vspace{-16pt}
\caption{\small{(a) The framework of our method. Our goal is to generate some noise $\mathcal{P}$ to disturb the input images $\mathcal{I}$. The disturbed images $\hat{\mathcal{I}}$ is able to cheat the ReID system $\mathcal{T}$ by attacking the visual similarities. (b) Specifically, the distance of each pair of samples from different categories (e.g., ($\hat{I}^k_c$, $I$), $\forall I\in\left\{I_{cd}\right\}$) is minimized, while the distance of each pair of the samples from the same category (e.g., ($\hat{I}^k_c$, $I$), $\forall I\in\left\{I_{cs}\right\}$) is maximized. The overall framework is trained by a generative adversarial network ($GAN$).}}\label{fig:method}
\vspace{-16pt}
\end{figure*}

\section{Related Work}

\noindent
\textbf{Person Re-identification.} ReID is different from image classification tasks in the setup of training and testing data. In an image classification task, the training and test set share the same categories, while in ReID, there is no category overlap between them.
Therefore,
deep ranking \cite{ding2015deep} is usually in desire for ReID.
However, deep ranking is sensitive to alignment. To address the (dis)alignment problem, several methods have been proposed by using structural messages \cite{sun2018beyond,li2018harmonious}.
Recently,
Zhang \textit{et al.}\cite{DBLP:journals/corr/abs-1711-08184} introduce the shortest path loss to supervise local parts aligning and adopt a mutual learning approach in the metric learning setting,
which has obtained the surpassing human-level performance.
Besides the supervised learning mentioned above, recent advance GANs have been introduced to ReID to boost performance in some unsupervised manner \cite{zheng2017unlabeled,zhong2018camera,zhong2018generalizing,deng2018image}.
Despite their success, the security and robustness of the existing ReID system have not yet been examined. Analyzing the robustness of a ReID system to resist attacks should be raised on the agenda.

\textbf{Adversarial Attacks.}
Since the discovery of adversarial examples for DNNs \cite{DBLP:journals/corr/SzegedyZSBEGF13}, several adversarial attacks have been proposed in recent years. Goodfellow \emph{et al.} \cite{Goodfellow2014ExplainingAH} proposes to generate adversarial examples by using a single step based on the sign of the gradient for each pixel, which often leads to sub-optimal results and the lack of generalization capacity.
Although DeepFool \cite{moosavi2016deepfool} is capable of fooling deep classifiers, it also lacks generalization capacity. Both methods fail to control the number of pixels to be attacked. To address this problem, \cite{papernot2016limitations} utilize the Jacobian matrix to implicitly conduct a fixed length of noise through the direction of each axis. Unfortunately, it cannot arbitrarily decide the number of target pixels to be attacked. \cite{su2019one} proposes to modify the single-pixel adversarial attack. However, the searching space and time grow dramatically with the increment of target pixels to be attacked. Besides the image classification, the adversarial attack is also introduced to face recognition \cite{Sharif2016Accessorize_SIGSAC,Dong2019Efficient_cvpr}. As discussed Section \ref{sect:intro}, all of the above methods do not fit the deep ranking problem. Also, their transferability is poor. Furthermore, many of them do not focus on the inconspicuousness of the visual quality. These drawbacks limit their applications in open-set tasks, e.g., ReID, which is our focus in this work. Although \cite{zhoumetric} has studied in metric analysis in person ReID, it does not provide a new adversarial attack method for ReID. It just uses the off-the-shelf methods for misclassification to examine very few ReID methods.

\section{Methodology}

\subsection{Overall Framework} \label{sect:overview}
The overall framework of our method is presented in Fig. \ref{fig:method} (a). Our goal is to use the generator $\mathcal{G}$ to produce deceptive noises $\mathcal{P}$ for each input image $\mathcal{I}$. By adding the noises $\mathcal{P}$ to the image $\mathcal{I}$, we obtain the adversarial example $\mathcal{\hat{I}}$, using which we are able to cheat the ReID system $\mathcal{T}$ to output the wrong results. Specifically, the ReID system $\mathcal{T}$ may consider the matched pair of images dissimilar, while considering the mismatched pair of images similar, as shown in Fig.\ref{fig:method} (b). The overall framework is trained by a generative adversarial network ($GAN$) with a generator $\mathcal{G}$ and a novel discriminator $\mathcal{D}$, which will be described in Section \ref{sect:transferable}.

\subsection{Learning-to-Mis-Rank Formulation For ReID}\label{sect:formulation}
We propose a learning-to-mis-rank formulation to perturb the ranking of system output. A new mis-ranking loss function is designed to attack the ranking of the predictions, which fits the ReID problem perfectly. Our method tends to minimize the distance of the mismatched pair and maximize the distance of the matched pair simultaneously. We have:\begin{small}\begin{equation}\label{eq:etri}
\begin{aligned}
 \mathcal{L}_{adv{\_}etri} = &\displaystyle\sum_{k=1}^{K}\sum_{c=1}^{C_k}\Big[\max_{\substack{j\ne k\\ j=1\dotso K\\ c_{d}=1\dotso C_j}}\big\|\mathcal{T}(\mathcal{\hat{I}}_c^k) - \mathcal{T}(\mathcal{\hat{I}}_{c_{d}}^j)\big\|_2^2\\
     & -\min_{\substack{c_{s}=1\dotso C_k}}\big\|\mathcal{T}(\mathcal{\hat{I}}_c^k)-\mathcal{T}(\mathcal{\hat{I}}_{c_{s}}^k)\big\|_2^2 + \Delta\Big]_{+},
\end{aligned}\end{equation}\end{small}where $C_k$ is the number of samples drawn from the k-$th$ person ID, $\mathcal{I}_c^k$ is the $c$-th images of the $k$ ID in a mini-batch, $c_s$ and $c_d$ are the samples from the same ID and the different IDs, $\big\|\cdot\big\|_2^2$ is the square of L2 norm used as the distance metric, and $\Delta$\ is a margin threshold. Eqn.\ref{eq:etri} attacks the deep ranking in the form of triplet loss \cite{ding2015deep}, where the distance of the \emph{easiest} distinguished pairs of inter-ID images are encouraged to small, while the distance of the \emph{easiest} distinguished pairs of intra-ID images are encouraged to large.

Remarkably, using the mis-ranking loss has a couple of advantages. \textbf{First}, the mis-ranking loss fits the ReID problem perfectly. As is mentioned above, ReID is different from image classification tasks in the setup of training and testing data. In an image classification task, the training and test set share the same categories, while in ReID, there is no category overlap between them. Therefore, the mis-ranking loss is suitable for attacking ReID. \textbf{Second}, the mis-ranking loss not only fits the ReID problem; it may fit all the open-set problems. Therefore, the use of mis-ranking loss may also benefit the learning of general and transferable features for the attackers. In summary, our mis-ranking based adversarial attacker is perfectly complementary to the existing misclassification based attackers.

\subsection{Learning Transferable Features for Attacking}\label{sect:transferable}
\begin{figure}[t]
\centering
\includegraphics[width=0.45\textwidth]{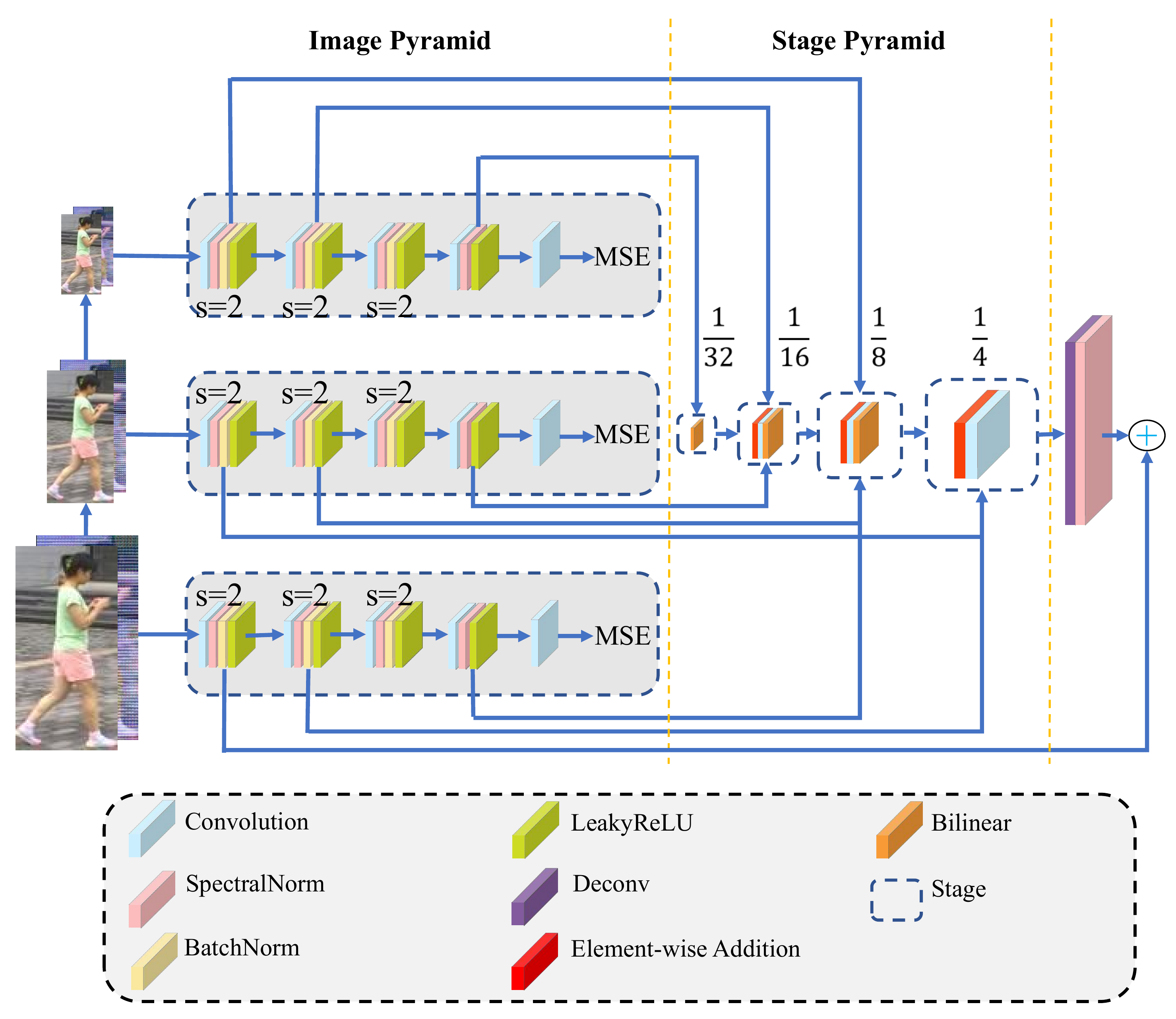}
\caption{\small{Detail of our multi-stage discriminator.
}}\label{fig:D}
\vspace{-11pt}
\end{figure}

As is suggested by \cite{Ilyas2019Adversarial_NIPS}, adversarial examples are features rather than bugs. Hence, to enhance the transferability of an attacker, one needs to improve the representation learning ability of the attacker to extract the general features for the adversarial perturbations. In our case, the representation learners are the generator $\mathcal{G}$ and the discriminator $\mathcal{D}$ (see Fig. \ref{fig:method} (a)). For the generator $\mathcal{G}$, we use the ResNet50. For the discriminator $\mathcal{D}$, recent adversarial defenders have utilized cross-layer information to identify adversarial examples\cite{liao2018defense,carlini2017towards,DBLP:conf/iclr/MetzenGFB17,li2017adversarial,xie2019feature}. As their rival, we develop a novel multi-stage network architecture for representation learning by pyramiding the features of different levels of the discriminator. Specifically, as shown in Fig. \ref{fig:D}, our discriminator $\mathcal{D}$ consists of three fully convolutional sub-networks, each of which includes five convolutional, three downsampling, and several normalization layers\cite{DBLP:conf/icml/IoffeS15,DBLP:conf/iclr/MiyatoKKY18}. The three sub-networks receives $\{1, 1/2^2, 1/4^2\}$ areas of the original images as the input, respectively. Next, the feature maps from these sub-networks with the same size are combined into the same \emph{stage} following \cite{lin2017feature}. A \emph{stage pyramid} with series of downsampled results with a ratio of $\{1/32, 1/16, 1/8, 1/4\}$ of the image is thus formulated. With the feature maps from the previous stage, we upsample the spatial resolution by a factor of 2 using bilinear upsampling and attach a $1\times 1$ convolutional layer to reduce channel dimensions. After an element-wise addition and a $3\times 3$ convolutions, the fused maps are fed into the next stage. Lastly, the network ends with two atrous convolution layers and a $1 \times 1$ convolution to perform feature re-weighting, whose final response map $\lambda$ is then fed into downstream sampler $\mathcal{M}$ discussed in Section \ref{sect:control}. Remarkably, all these three sub-networks are optimized by standard loss following \cite{mao2017least}.

\subsection{Controlling the Number of the Attacked Pixels}\label{sect:control}
To make our attack inconspicuous, we improve the existing attackers in two aspects. The first aspect is to control the number of the target pixels to be attacked. Generally, an adversarial attack is to introduce a set of noise to a set of target pixels for a given image to form an adversarial example. Both the noise and the target pixels are unknown, which will be searched by the attacker. Here, we present the formulation of our attacker in searching for the target pixels. To make the search space continuous, we relax the choice of a pixel as a Gumbel softmax over all possible pixels:\begin{small}\begin{equation}
p_{i,j} = \frac{\exp((log(\lambda_{i,j}+\mathcal{N}_{i,j}))/\tau)}{\sum_{i,j=1}^{H,W}\exp(log(\lambda_{i,j}+\mathcal{N}_{i,j})/\tau)},
\end{equation}\end{small}where $i\in(0,H),j\in(0,W)$ denote the index of pixel in a feature map of size $H\times W$, where $H$/$W$ are the height/width of the input images. The probability $p_{i,j}$ of a pixel to be chosen is parameterized by a softmax output vector $\lambda_{i,j}$ of dimension $H\times W$. $\mathcal{N}_{i,j}=-log(-log(U))$ is random variable at position $(i,j)$, which is sampled from Gumbel distribution\cite{gumbel1954statistical} with $U\sim Uniform(0,1)$. Note that $\tau$ is a temperature parameter to soften transition from uniform distribution to categorical distribution when $\tau$ gradually reduces to zero. Thus, the number of the target pixels to be attacked is determined by the mask $\mathcal{M}:$\begin{small}\begin{equation}\label{eqw:mask}
\mathcal{M}_{ij}=
\left\{
\begin{aligned}
&\mathcal{K}eep\mathcal{T}opk(p_{i,j}), &\text{in forward propagation}\\
&p_{i,j}, &\text{in backward propagation}
\end{aligned}
\right.
\end{equation}\end{small}where $\mathcal{K}eep\mathcal{T}opk$ is a function by which the top-k pixels with the highest probability $p_{i,j}$ are retained in $\mathcal{M}$ while the other pixels are dropped during the forward propagation. Moreover, the difference between the forward and backward propagation ensures the differentiability. By multiplying the mask $\mathcal{M}$ and the preliminary noise $\mathcal{P}'$, we obtain the final noise $\mathcal{P}$ with controllable number of activated pixels. The usage of $\mathcal{M}$ is detailed in Fig. \ref{fig:method} (a).

\subsection{Perception Loss for Visual Quality}\label{sect:detail}
In addition to controlling the number of the attacked pixels, we also focus on the visual quality to ensure the inconspicuousness of our attackers. Existing works introduce noises to images to cheat the machines without considering the visual quality of the images, which is inconsistent with human cognition. Motivated by MS-SSIM \cite{wang2003multiscale} that is able to provide a good approximation to perceive image quality for visual perception, we include an perception loss $\mathcal{L}_{VP}$ in our formulation to improve the visual quality:\begin{small}\begin{equation}\label{eqn:SSIM}
\mathcal{L}_{VP}(\mathcal{I}, \mathcal{\hat{I}}) = [l_L(\mathcal{I}, \mathcal{\hat{I}})]^{\alpha_L} \cdot \displaystyle\prod_{j=1}^{L}[c_j(\mathcal{I}, \mathcal{\hat{I}})]^{\beta_j}[s_j(\mathcal{I}, \mathcal{\hat{I}})]^{\gamma_j},
\end{equation}\end{small}
where $c_j$ and $s_j$ are the measures of the contrast comparison and the structure comparison at the $j$-th scale respectively, which are calculated by \begin{small}$c_j(\mathcal{I}, \hat{\mathcal{I}})=\frac{2 \sigma_{\mathcal{I}} \sigma_{\hat{\mathcal{I}}}+C_{2}}{\sigma_{\mathcal{I}}^{2}+\sigma_{\hat{\mathcal{I}}}^{2}+C_{2}}$\end{small} and \begin{small}$s_j(\mathcal{I}, \hat{\mathcal{I}})=\frac{\sigma_{\mathcal{I} \hat{\mathcal{I}}}+C_{3}}{\sigma_{\mathcal{I}} \sigma_{\hat{\mathcal{I}}}+C_{3}}$\end{small}, where $\sigma$ is the variance/covariance. $L$ is the level of scales, $\alpha_L$, $\beta_j$, and $\gamma_j$ are the factors to re-weight the contribution of each component. Thanks to $\mathcal{L}_{VP}$, the attack with high magnitude is available without being noticed by humans.

\subsection{Objective Function}

Besides the mis-ranking loss $\mathcal{L}_{adv{\_}etri}$, the perception loss $\mathcal{L}_{VP}$, we have two additional losses, i.e., a misclassification loss $\mathcal{L}_{adv{\_}xent}$, and a GAN loss $\mathcal{L}_{GAN}$.

\vspace{3pt}
{\textbf{Misclassification Loss.}} Existing works usually consider the least likely class as the target to optimize the cross-entropy between the output probabilities and its least likely class. However, the model may misclassify the inputs as any class except for the correct one. Inspired by \cite{szegedy2016rethinking}, we propose a mechanism for relaxing the model for non-targeted attack by:\begin{small}\begin{equation}
\label{eqn:adv_xent}
\mathcal{L}_{adv{\_}xent} = -\displaystyle\sum_{k=1}^{K}\mathcal{S}(\mathcal{T}(\mathcal{\hat{I}}))_k((1-\delta)\mathbbm{1}_{\arg\min{\mathcal{T}(\mathcal{I})_k}}+\delta v_k),
\end{equation}\end{small}where $\mathcal{S}$ denotes the log-softmax function, $K$ is the total number of person IDs and \begin{small}$v=[\frac{1}{K-1},\dotsc , 0 ,\dotsc ,\frac{1}{K-1}]$\end{small} is smoothing regularization in which $v_k$ equals to \begin{small}$\frac{1}{K-1}$\end{small} everywhere except when $k$ is the ground-truth ID. The term $\arg\min$ in Eqn. \ref{eqn:adv_xent} is similar to numpy.argmin which returns the indices of the minimum values of an output probability vector, indicating the least likely class. In practice, this smoothing regularization improves the training stability and the success attack rate.

\renewcommand{\arraystretch}{1.1}
\begin{table*}[!htb]
\centering
\tiny
\caption{\small{Attacking the state-of-the-art ReID systems. \emph{IDE:} \cite{DBLP:journals/corr/ZhengYH16}; \emph{DenseNet-121:} \cite{huang2017densely}; \emph{Mudeep:} \cite{fu2017multi}; \emph{AlignedReid:} \cite{DBLP:journals/corr/abs-1711-08184}; \emph{PCB:} \cite{sun2018beyond}; \emph{HACNN:} \cite{li2018harmonious}; \emph{LSRO:} \cite{zheng2017unlabeled}; \emph{HHL:} \cite{zhong2018generalizing}; \emph{SPGAN:} \cite{deng2018image}; \emph{CamStyle+Era:} \cite{zhong2018camera}. We select GAP\cite{Poursaeed_2018_CVPR} and PGD\cite{DBLP:conf/iclr/MadryMSTV18} as the baseline attackers.}}\label{tab:big}
\begin{tabular}{clcccccccccccccccc}
\multicolumn{18}{c}{(a) Market1501} \\ \hline
\multicolumn{2}{c|}{\multirow{2}{*}{Methods}} & \multicolumn{4}{c|}{Rank-1} & \multicolumn{4}{c|}{Rank-5} & \multicolumn{4}{c|}{Rank-10} & \multicolumn{4}{c}{mAP} \\ \cline{3-18}
\multicolumn{2}{c|}{} & Before & \multicolumn{1}{l}{GAP} & \multicolumn{1}{l}{PGD} & \multicolumn{1}{c|}{Ours} & Before & \multicolumn{1}{l}{GAP} & \multicolumn{1}{l}{PGD} & \multicolumn{1}{c|}{Ours} & Before & \multicolumn{1}{l}{GAP} & \multicolumn{1}{l}{PGD} & \multicolumn{1}{c|}{Ours} & Before & \multicolumn{1}{l}{GAP} & \multicolumn{1}{l}{PGD} & Ours \\ \hline
\multicolumn{1}{c|}{\multirow{3}{*}{Backbone}} & \multicolumn{1}{l|}{IDE (ResNet-50)} & 83.1 & 5.0 & 4.5 & \multicolumn{1}{c|}{\textbf{3.7}} & 91.7 & 10.0 & 8.7 & \multicolumn{1}{c|}{\textbf{8.3}} & 94.6 & 13.9 & 12.1 & \multicolumn{1}{c|}{\textbf{11.5}} & 63.3 & 5.0 & 4.6 & \textbf{4.4} \\
\multicolumn{1}{c|}{} & \multicolumn{1}{l|}{DenseNet-121} & 89.9 & 2.7 & \textbf{1.2} & \multicolumn{1}{c|}{\textbf{1.2}} & 96.0 & 6.7 & \textbf{1.0} & \multicolumn{1}{c|}{1.3} & 97.3 & 8.5 & \textbf{1.5} & \multicolumn{1}{c|}{2.1} & 73.7 & 3.7 & \textbf{1.3} & \textbf{1.3} \\
\multicolumn{1}{c|}{} & \multicolumn{1}{l|}{Mudeep (Inception-V3)} & 73.0 & 3.5 & 2.6 & \multicolumn{1}{c|}{\textbf{1.7}} & 90.1 & 5.3 & 5.5 & \multicolumn{1}{c|}{\textbf{1.7}} & 93.1 & 7.6 & 6.9 & \multicolumn{1}{c|}{\textbf{5.0}} & 49.9 & 2.8 & 2.0 & \textbf{1.8} \\ \hline
\multicolumn{1}{c|}{\multirow{3}{*}{Part-Aligned}} & \multicolumn{1}{l|}{AlignedReid} & 91.8 & 10.1 & 10.2 & \multicolumn{1}{c|}{\textbf{1.4}} & 97.0 & 18.7 & 15.8 & \multicolumn{1}{c|}{\textbf{3.7}} & 98.1 & 23.2 & 19.1 & \multicolumn{1}{c|}{\textbf{5.4}} & 79.1 & 9.7 & 8.9 & \textbf{2.3} \\
\multicolumn{1}{c|}{} & \multicolumn{1}{l|}{PCB} & 88.6 & 6.8 & 6.1 & \multicolumn{1}{c|}{\textbf{5.0}} & 95.5 & 14.0 & 12.7 & \multicolumn{1}{c|}{\textbf{10.7}} & 97.3 & 19.2 & 15.8 & \multicolumn{1}{c|}{\textbf{14.3}} & 70.7 & 5.6 & 4.8 & \textbf{4.3} \\
\multicolumn{1}{c|}{} & \multicolumn{1}{l|}{HACNN} & 90.6 & 2.3 & 6.1 & \multicolumn{1}{c|}{\textbf{0.9}} & 95.9 & 5.2 & 8.8 & \multicolumn{1}{c|}{\textbf{1.4}} & 97.4 & 6.9 & 10.6 & \multicolumn{1}{c|}{\textbf{2.3}} & 75.3 & 3.0 & 5.3 & \textbf{1.5} \\ \hline
\multicolumn{1}{l|}{\multirow{4}{*}{Data Augmentation}} & \multicolumn{1}{l|}{CamStyle+Era (IDE)} & 86.6 & 6.9 & 15.4 & \multicolumn{1}{c|}{\textbf{3.9}} & 95.0 & 14.1 & 23.9 & \multicolumn{1}{c|}{\textbf{7.5}} & 96.6 & 18.0 & 29.1 & \multicolumn{1}{c|}{\textbf{10.0}} & 70.8 & 6.3 & 12.6 & \textbf{4.2} \\
\multicolumn{1}{l|}{} & \multicolumn{1}{l|}{LSRO (DenseNet-121)} & 89.9 & 5.0 & 7.2 & \multicolumn{1}{c|}{\textbf{0.9}} & 96.1 & 10.2 & 13.1 & \multicolumn{1}{c|}{\textbf{2.2}} & 97.4 & 12.6 & 15.2 & \multicolumn{1}{c|}{\textbf{3.1}} & 77.2 & 5.0 & 8.1 & \textbf{1.3} \\
\multicolumn{1}{l|}{} & \multicolumn{1}{l|}{HHL (IDE)} & 82.3 & 5.0 & 5.7 & \multicolumn{1}{c|}{\textbf{3.6}} & 92.6 & 9.8 & 9.8 & \multicolumn{1}{c|}{\textbf{7.3}} & 95.4 & 13.5 & 12.2 & \multicolumn{1}{c|}{\textbf{9.7}} & 64.3 & 5.4 & 5.5 & \textbf{4.1} \\
\multicolumn{1}{l|}{} & \multicolumn{1}{l|}{SPGAN (IDE)} & 84.3 & 8.8 & 10.1 & \multicolumn{1}{c|}{\textbf{1.5}} & 94.1 & 18.6 & 16.7 & \multicolumn{1}{c|}{\textbf{3.1}} & 96.4 & 24.5 & 20.9 & \multicolumn{1}{c|}{\textbf{4.3}} & 66.6 & 8.0 & 8.6 & \textbf{1.6} \\ \hline
\multicolumn{18}{c}{(b) CUHK03} \\ \hline
\multicolumn{2}{c|}{\multirow{2}{*}{Methods}} & \multicolumn{4}{c|}{Rank-1} & \multicolumn{4}{c|}{Rank-5} & \multicolumn{4}{c|}{Rank-10} & \multicolumn{4}{c}{mAP} \\ \cline{3-18}
\multicolumn{2}{c|}{} & Before & \multicolumn{1}{l}{GAP} & \multicolumn{1}{l}{PGD} & \multicolumn{1}{c|}{Ours} & Before & \multicolumn{1}{l}{GAP} & \multicolumn{1}{l}{PGD} & \multicolumn{1}{c|}{Ours} & Before & \multicolumn{1}{l}{GAP} & \multicolumn{1}{l}{PGD} & \multicolumn{1}{c|}{Ours} & Before & \multicolumn{1}{l}{GAP} & \multicolumn{1}{l}{PGD} & Ours \\ \hline
\multicolumn{1}{c|}{\multirow{3}{*}{Backbone}} & \multicolumn{1}{l|}{IDE (ResNet-50)} & 24.9 & 0.9 & 0.8 & \multicolumn{1}{c|}{\textbf{0.4}} & 43.3 & 2.0 & 1.2 & \multicolumn{1}{c|}{\textbf{0.7}} & 51.8 & 2.9 & 2.1 & \multicolumn{1}{c|}{\textbf{1.5}} & 24.5 & 1.3 & \textbf{0.8} & \textbf{0.9} \\
\multicolumn{1}{c|}{} & \multicolumn{1}{l|}{DenseNet-121} & 48.4 & 2.4 & 0.1 & \multicolumn{1}{c|}{\textbf{0.0}} & 50.1 & 4.4 & \textbf{0.1} & \multicolumn{1}{c|}{\textbf{0.2}} & 70.1 & 5.9 & \textbf{0.3} & \multicolumn{1}{c|}{0.6} & 84.0 & 1.6 & \textbf{0.2} & \textbf{0.3} \\
\multicolumn{1}{c|}{} & \multicolumn{1}{l|}{Mudeep (Inception-V3)} & 32.1 & 1.1 & 0.4 & \multicolumn{1}{c|}{\textbf{0.1}} & 53.3 & 3.7 & 1.0 & \multicolumn{1}{c|}{\textbf{0.5}} & 64.1 & 5.6 & 1.5 & \multicolumn{1}{c|}{\textbf{0.8}} & 30.1 & 2.0 & 0.8 & \textbf{0.3} \\ \hline
\multicolumn{1}{c|}{\multirow{3}{*}{Part-Aligned}} & \multicolumn{1}{l|}{AlignedReid} & 61.5 & 2.1 & \textbf{1.4} & \multicolumn{1}{c|}{\textbf{1.4}} & 79.4 & 4.6 & \textbf{2.2} & \multicolumn{1}{c|}{3.7} & 85.5 & 6.2 & \textbf{4.1} & \multicolumn{1}{c|}{5.4} & 59.6 & 3.4 & \textbf{2.1} & \textbf{2.1} \\
\multicolumn{1}{c|}{} & \multicolumn{1}{l|}{PCB} & 50.6 & 0.9 & 0.5 & \multicolumn{1}{c|}{\textbf{0.2}} & 71.4 & 4.5 & 2.1 & \multicolumn{1}{c|}{\textbf{1.3}} & 78.7 & 5.8 & 4.5 & \multicolumn{1}{c|}{\textbf{1.8}} & 48.6 & 1.4 & 1.2 & \textbf{0.8} \\
\multicolumn{1}{c|}{} & \multicolumn{1}{l|}{HACNN} & 48.0 & 0.9 & 0.4 & \multicolumn{1}{c|}{\textbf{0.1}} & 69.0 & 2.4 & 0.9 & \multicolumn{1}{c|}{\textbf{0.3}} & 78.1 & 3.4 & 1.3 & \multicolumn{1}{c|}{\textbf{0.4}} & 47.6 & 1.8 & 0.8 & \textbf{0.4} \\ \hline
\multicolumn{18}{c}{(c) DukeMTMC} \\ \hline
\multicolumn{2}{c|}{\multirow{2}{*}{Methods}} & \multicolumn{4}{c|}{Rank-1} & \multicolumn{4}{c|}{Rank-5} & \multicolumn{4}{c|}{Rank-10} & \multicolumn{4}{c}{mAP} \\ \cline{3-18}
\multicolumn{2}{c|}{} & Before & GAP & \multicolumn{1}{l}{PGD} & \multicolumn{1}{c|}{Ours} & Before & GAP & \multicolumn{1}{l}{PGD} & \multicolumn{1}{c|}{Ours} & Before & GAP & \multicolumn{1}{l}{PGD} & \multicolumn{1}{c|}{Ours} & Before & GAP & \multicolumn{1}{l}{PGD} & Ours \\ \hline
\multicolumn{1}{c|}{\multirow{4}{*}{Data augmentation}} & \multicolumn{1}{l|}{CamStyle+Era (IDE)} & 76.5 & 3.3 & 22.9 & \multicolumn{1}{c|}{\textbf{1.2}} & 86.8 & 7.0 & 34.1 & \multicolumn{1}{c|}{\textbf{2.6}} & 90.0 & 9.6 & 39.9 & \multicolumn{1}{c|}{\textbf{3.4}} & 58.1 & 3.5 & 16.8 & \textbf{1.5} \\
\multicolumn{1}{c|}{} & \multicolumn{1}{l|}{LSRO (DenseNet-121)} & 72.0 & 1.3 & 7.2 & \multicolumn{1}{c|}{\textbf{0.7}} & 85.7 & 2.9 & 12.5 & \multicolumn{1}{c|}{\textbf{1.6}} & 89.5 & 4.0 & 18.4 & \multicolumn{1}{c|}{\textbf{2.2}} & 55.2 & 1.4 & 8.1 & \textbf{0.9} \\
\multicolumn{1}{c|}{} & \multicolumn{1}{l|}{HHL (IDE)} & 71.4 & 1.8 & 9.5 & \multicolumn{1}{c|}{\textbf{1.0}} & 83.5 & 3.4 & 15.6 & \multicolumn{1}{c|}{\textbf{2.0}} & 87.7 & 4.2 & 19.0 & \multicolumn{1}{c|}{\textbf{2.5}} & 51.8 & 1.9 & 7.4 & \textbf{1.3} \\
\multicolumn{1}{c|}{} & \multicolumn{1}{l|}{SPGAN (IDE)} & 73.6 & 5.3 & 12.4 & \multicolumn{1}{c|}{\textbf{0.1}} & 85.2 & 10.3 & 21.1 & \multicolumn{1}{c|}{\textbf{0.5}} & 88.9 & 13.4 & 26.3 & \multicolumn{1}{c|}{\textbf{0.6}} & 54.6 & 4.7 & 10.2 & \textbf{0.3} \\ \hline
\end{tabular}
\vspace{-11pt}
\end{table*}

\vspace{3pt}
{\textbf{GAN Loss.}} For our task, the generator $\mathcal{G}$ attempts to produce deceptive noises from input images, while the discriminator D distinguishes real images from adversarial examples as much as possible. Hence, the GAN loss $\mathcal{L}_{GAN}$ is given as:\begin{small}\begin{equation}\label{eqn:GAN}
\mathcal{L}_{GAN} = \mathbb{E}_{(I_{cd},I_{cs})} [\log \mathcal{D}_{1,2,3}(I_{cd},I_{cs})] + \mathbb{E}_{\mathcal{I}} [\log(1 - \mathcal{D}_{1,2,3}(\mathcal{I}, \mathcal{\hat{I}}))],
\end{equation}\end{small}where $\mathcal{D}_{1,2,3}$ is our multi-stage discriminator shown in Fig. \ref{fig:D}.
We access to the final loss function:\begin{small}\begin{equation}\label{eqn:full}
\mathcal{L} = \mathcal{L}_{GAN} + \mathcal{L}_{adv{\_}xent} + \zeta\mathcal{L}_{adv{\_}etri} + \eta(1-\mathcal{L}_{VP}),
\end{equation}\end{small}where $\zeta$ and $\eta$ are loss weights for balance.
\section{Experiment}
We first present the results of attacking state-of-the-art ReID systems and then perform ablation studies on our method. Then, the generalization ability and interpretability of our method are examined by exploring black-box attacks.

\textbf{Datasets.} Our method is evaluated on four of the largest ReID datasets: Market1501 \cite{zheng2015scalable}, CUHK03 \cite{li2014deepreid} DukeMTMC \cite{ristani2016performance} and MSMT17 \cite{wei2018person}. Market1501 is a fully studied dataset containing 1,501 identities and 32,688 bounding boxes. CUHK03 includes 1,467 identities and 28,192 bounding boxes. To be consistent with recent works, we follow the \textbf{new} training/testing protocol to perform our experiments \cite{zhong2017re}. DukeMTMC provides 16,522 bounding boxes of 702 identities for training and 17,661 for testing. MSMT17 covers 4,101 identities and 126,441 bounding boxes taken by 15 cameras in both indoor and outdoor scenes. We adopt the standard metric of mAP and rank-$\{1, 5, 10, 20\}$ accuracy for evaluation. \emph{Note that in contrast to a ReID problem, lower rank accuracy and mAP indicate better success attack rate in a attack problem.}

\textbf{Protocols.} The details about training protocols and hyper-parameters can be seen in Appendix \textcolor{red}{C}. The first two subsections validate a white-box attack, i.e., the attacker has full access to training data and target models. In the third subsection, we explore a black-box attack to examine the transferability and interpretability of our method, i.e., the attacker has no access to the training data and target models. Following the standard protocols of the literature, all experiments below are performed by $L_\infty$-bounded attacks with $\varepsilon=16$ without special instruction, where $\varepsilon$ is an upper bound imposed on the amplitude of the generated noise (\{\begin{scriptsize}$\|\mathcal{\hat{I}}-\mathcal{I}\|_{1,2,\operatorname{or}\infty} \leq \epsilon$\end{scriptsize}\}) that determines the attack intensity and the visual quality.

\subsection{Attacking State-of-the-Art ReID Systems}\label{sect:sota}
\noindent
To demonstrate the generality of our method, we divide the state-of-the-art ReID systems into three groups as follows.

\textbf{Attacking Different Backbones.} We first examine the effectiveness of our method in attacking different best-performing network backbones, including ResNet-50 \cite{he2016deep} (i.e., IDE \cite{DBLP:journals/corr/ZhengYH16}), DenseNet-121 \cite{huang2017densely}, and Inception-v3 \cite{szegedy2016rethinking} (i.e., Mudeep \cite{fu2017multi}). The results are shown in Table \ref{tab:big} (a) \& (b). We can see that the rank-1 accuracy of all backbones drop sharply approaching zero (e.g, from 89.9\% to 1.2\% for DenseNet) after it has been attacked by our method, suggesting that changing backbones cannot defend our attack.

\textbf{Attacking Part-based ReID Systems.} Many best-performing ReID systems learn both local and global similarity by considering part alignment. However, they still fail to defend our attack (Table \ref{tab:big} (a)(b)). For example, the accuracy of one of the best-performing ReID systems (AlignedReID \cite{DBLP:journals/corr/abs-1711-08184}) drops sharply from 91.8\% to 1.4\% after it has been attacked by our method. This comparison proves that the testing tricks, e.g., extra local features ensemble in AlignedReID \cite{DBLP:journals/corr/abs-1711-08184} and flipped image ensembling in PCB \cite{sun2018beyond}, are unable to resist our attack.

\textbf{Attacking Augmented ReID Systems.} Many state-of-the-art ReID systems use the trick of data augmentation. Next, we examine the effectiveness of our model in attacking these augmentation-based systems. Rather than conventional data augmentation trick (e.g., random cropping, flipping, and 2D-translation), we examine four new augmentation tricks using GAN to increase the training data. The evaluation is conducted on Market1501 and DukeMTMC. The results in Table \ref{tab:big} (a)(c) show that although GAN data augmentations improve the ReID accuracy, they cannot defend our attack. In contrast, we have even observed that better ReID accuracy may lead to worse robustness.

\emph{\textbf{Discussion.} We have three remarks for rethinking the robustness of ReID systems for future improvement. {\textbf{First}}, there is no effective way so far to defend against our attacks, e.g., after our attack, all rank-1 accuracies drop below 3.9\%. {\textbf{Second}}, the robustness of Mudeep \cite{fu2017multi} and PCB \cite{sun2018beyond} are strongest. Intuitively, Mudeep may benefit from its nonlinear and large receptive field. For PCB, reprocessing the query images and hiding the network architecture during evaluation may improve the robustness. {\textbf{Third}}, HACNN \cite{li2018harmonious} has the lowest rank-1 to rank-20 accuracy after the attack, suggesting that attention mechanism may be harmful to the defensibility. The returns from the target ReID system before and after the adversarial attack are provided in Appendix \textcolor{red}{A}.}

\begin{table*}[!htb]
\centering
\caption{\small{\textbf{Ablations.} We present six major ablation experiments in this table. \textbf{R-1,R-5,\& R-10:} Rank-1, Rank-5, \& Rank-10.}}\label{tab:ablation}
\scriptsize
\begin{tabular}{p{0.005cm}cp{0.3691cm}cp{0.005cm}cp{0.005cm}cp{0.3691cm}cp{0.005cm}cp{0.005cm}cp{0.005cm}cp{0.005cm}cp{0.005cm}cp{0.005cm}cp{0.005cm}cp{0.005cm}cp{0.005cm}cp{0.005cm}cp{0.005cm}cp{0.005cm}p{0.005cm}c}
\multicolumn{1}{l|}{} & R-1 & R-5 & R-10 & mAP & \multicolumn{1}{c}{} & \multicolumn{1}{l|}{} & R-1 & R-5 & R-10 & mAP & \multicolumn{1}{c}{} & \multicolumn{1}{l|}{} & \multicolumn{1}{c}{R-1} & \multicolumn{1}{c}{R-5} & \multicolumn{1}{c}{R-10} & \multicolumn{1}{c}{mAP} \\ \clineB{1-5}{3.0} \clineB{7-11}{3.0} \clineB{13-17}{3.0}
\multicolumn{1}{l|}{(A) cent} & 28.5 & 43.9 & 51.4 & 23.8 & & \multicolumn{1}{l|}{$\epsilon$=$40$} & 0.0 & 0.2 & 0.6 & 0.2 & & \multicolumn{1}{l|}{PatchGAN ($\epsilon$=$40$)} & \multicolumn{1}{c}{48.3} & \multicolumn{1}{c}{65.8} & \multicolumn{1}{c}{73.1} & \multicolumn{1}{c}{37.7} \\
\multicolumn{1}{l|}{(B) xent} & 13.7 & 22.5 & 28.7 & 12.5 & & \multicolumn{1}{l|}{$\epsilon$=$20$} & 0.1 & 0.4 & 0.8 & 0.4 & & \multicolumn{1}{l|}{Ours ($\epsilon$=$40$)} & \multicolumn{1}{c}{\textbf{0.0}} & \multicolumn{1}{c}{\textbf{0.2}} & \multicolumn{1}{c}{\textbf{0.6}} & \multicolumn{1}{c}{\textbf{0.2}} \\ \clineB{1-5}{3.0}
\multicolumn{1}{l|}{(C) etri} & \textbf{4.5} & 9.1 & 12.5 & \textbf{5.1} & & \multicolumn{1}{l|}{$\epsilon$=$16$} & 1.4 & 3.7 & 5.4 & 2.3 & & \multicolumn{1}{l|}{PatchGAN ($\epsilon$=$10$)} & \multicolumn{1}{c}{53.3} & \multicolumn{1}{c}{69.2} & \multicolumn{1}{c}{75.6} & \multicolumn{1}{c}{43.2} \\
\multicolumn{1}{l|}{(D) xent+etri} & \textbf{1.4} & \textbf{3.7} & \textbf{5.4} & \textbf{2.3} & & \multicolumn{1}{l|}{$\epsilon$=$10$} & 24.4 & 38.5 & 46.6 & 21.0 & & \multicolumn{1}{l|}{Ours ($\epsilon$=$10$)} & \multicolumn{1}{c}{\textbf{24.4}} & \multicolumn{1}{c}{\textbf{38.5}} & \multicolumn{1}{c}{\textbf{46.6}} & \multicolumn{1}{c}{\textbf{21.0}} \\
\multicolumn{5}{l}{\begin{tabular}[c]{@{}l@{}}(a) \textbf{Different Objectives}: The modified xent loss out-\\ performs the cent loss, but both of them are unstable.\\ Our loss brings more stable and higher fooling rate\\ than misclassification.\end{tabular}} & & \multicolumn{5}{l}{\begin{tabular}[c]{@{}l@{}}(b) \textbf{Comparisons of different $\epsilon$}: Results on \\ the variants of our model using different $\epsilon$.\\Our proposed method achieves good results\\even when $\epsilon=10$.\end{tabular}} & & \multicolumn{5}{l}{\begin{tabular}[c]{@{}l@{}}(c) \textbf{Multi-stage vs. Common discriminator}: Multi-\\stage technique improves results under both large\\and small $\epsilon$ for utilizing the information from\\previous layers.\end{tabular}} \\
                   & \multicolumn{1}{l}{} & \multicolumn{1}{l}{} & \multicolumn{1}{l}{} & \multicolumn{1}{l}{} & & & \multicolumn{1}{l}{} & \multicolumn{1}{l}{} & \multicolumn{1}{l}{} & \multicolumn{1}{l}{} & & & & & & \\
\multicolumn{1}{l|}{} & R-1 & R-5 & R-10 & mAP & & \multicolumn{1}{l|}{} & R-1 & R-5 & R-10 & mAP & & \multicolumn{1}{l|}{} & \multicolumn{1}{c}{R-1} & \multicolumn{1}{c}{R-5} & \multicolumn{1}{c}{R-10} & \multicolumn{1}{c}{mAP} \\ \clineB{1-5}{3.0} \clineB{7-11}{3.0} \clineB{13-17}{3.0}
\multicolumn{1}{l|}{Market$\rightarrow$CUHK} & 4.9 & 9.2 & 12.1 & 6.0 & & \multicolumn{1}{l|}{$\rightarrow$PCB} & 31.7 & 46.1 & 53.2 & 22.9 & & \multicolumn{1}{l|}{$\rightarrow$PCB(C)} & 6.9 & 12.9 & 18.9 & 8.2 \\
\multicolumn{1}{l|}{CUHK$\rightarrow$Market} & 34.3 & 51.6 & 58.6 & 28.2 & & \multicolumn{1}{l|}{$\rightarrow$HACNN} & 14.8 & 24.4 & 29.8 & 13.4 & & \multicolumn{1}{l|}{$\rightarrow$HACNN(C)} & 3.6 & 7.1 & 9.2 & 4.6 \\
\multicolumn{1}{l|}{Market$\rightarrow$Duke} & 17.7 & 26.7 & 32.6 & 14.2 & & \multicolumn{1}{l|}{$\rightarrow$LSRO} & 17.0 & 28.9 & 35.1 & 14.8 & & \multicolumn{1}{l|}{$\rightarrow$LSRO(D)} & 19.4 & 30.2 & 34.7 & 15.2 \\
\multicolumn{1}{l|}{Market$\rightarrow$MSMT} & 35.1 & 49.4 & 55.8 & 27.0 & & & \multicolumn{1}{l}{} & \multicolumn{1}{l}{} & \multicolumn{1}{l}{} & \multicolumn{1}{l}{} & & \multicolumn{1}{l|}{$\rightarrow$Mudeep(C)*} & 19.4 & 27.7 & 34.9 & 16.2 \\
\multicolumn{1}{l}{} & & & \multicolumn{1}{l}{} & \multicolumn{1}{l}{} & \multicolumn{1}{l}{} & \multicolumn{1}{l}{} & & & & & & \\
\multicolumn{5}{l}{\begin{tabular}[c]{@{}l@{}}(d) Crossing Dataset. \textbf{Market$\rightarrow$CUHK}: noises\\learned from Market1501 mislead inferring on\\ CUHK03. All experiments are based on Aligned-\\ ReID model.\end{tabular}} & & \multicolumn{5}{l}{\begin{tabular}[c]{@{}l@{}}(e) Crossing Model. \textbf{$\rightarrow$PCB}: noises learned\\from \textbf{AlignedReID} attack pretrained PCB\\model. All experiments are performed on\\Market1501.\end{tabular}} & & \multicolumn{5}{l}{\begin{tabular}[c]{@{}l@{}}(f) Crossing Dataset \& Model. \textbf{$\rightarrow$ PCB(C)}:\\noises learned from \textbf{AlignedReID} pretrained on\\Market-1501 are borrowed to attack PCB model\\inferred on CUHK03. \textbf{*} denotes \textbf{4k}-pixel attack.\end{tabular}} \\
                   & \multicolumn{1}{l}{} & \multicolumn{1}{l}{} & \multicolumn{1}{l}{} & \multicolumn{1}{l}{} & & & \multicolumn{1}{l}{} & \multicolumn{1}{l}{} & \multicolumn{1}{l}{} & \multicolumn{1}{l}{} & & & & & &             
\end{tabular}
\vspace{-22pt}
\end{table*}

\begin{table}[!ht]
\centering
\caption{\small{Proportion of adversarial points. $\dagger$ denotes the results with appropriate relaxation.}}\label{tab:num}
\scriptsize
\begin{tabular}{l|cccc}
   & R-1 & R-5 & R-10 & mAP \\ \hlineB{3.5}
Full size & 1.4 & 3.7 & 5.4 & 2.3 \\ \hlineB{2.5}
Ratio=1/2 & 39.3 & 55.0 & 62.4 & 31.5 \\
Ratio=1/4 & 72.7 & 85.9 & 89.7 & 58.3 \\
Ratio=1/4$\dagger$ & 0.3 & 1.5 & 2.7 & 0.7 \\
Ratio=1/8$\dagger$ & 0.6 & 1.8 & 3.0 & 1.1 \\
Ratio=1/16$\dagger$ & 8.2 & 14.7 & 17.8 & 6.9 \\
Ratio=1/32$\dagger$ & 59.4 & 76.5 & 82.2 & 47.3 \\
Ratio=1/64$\dagger$ & 75.5 & 87.6 & 91.6 & 61.5
\end{tabular}
\vspace{-11pt}
\end{table}

\subsection{Ablation Study}\label{sect:ablation}
\noindent
We conduct comprehensive studies to validate the effectiveness of each component of our method. AlignedReID \cite{DBLP:journals/corr/abs-1711-08184} is used as our target model in the rest of the paper for its remarkable results in ReID domain.

\textbf{Different Losses.} We report the rank-1 accuracy of four different losses to validate the effectiveness of our loss. The results are shown in Table \ref{tab:ablation} (a), where the four rows represent \textbf{(A)} the conventional misclassification loss, \textbf{(B)} our misclassification, \textbf{(C)} our mis-ranking loss, and \textbf{(D)} our misclassification + our mis-ranking loss, respectively. Actually, we observe that conventional misclassification loss A is incompatible with the perception loss, leading to poor attack performance (28.5\%). In contrast, our visual mis-ranking loss D achieves very appealing attack performance (1.4\%). We also observe that our misclassification loss B and our visual mis-ranking loss C benefit each other. Specifically, by combining these two losses, we obtain Loss D, which outperforms all the other losses.

\textbf{Multi-stage vs. Common Discriminator.} To validate the effectiveness of our multi-stage discriminator, we compare the following settings: \textbf{(A)} using our multi-stage discriminator and \textbf{(B)} using a commonly used discriminator. Specifically, we replace our multi-stage discriminator with PatchGAN\cite{isola2017image}. Table \ref{tab:ablation} (c) shows a significant degradation of attack performance after changing the discriminator, demonstrating the superiority of our multi-stage discriminator to capture more details for a better attack.

\begin{figure}
\centering
\includegraphics[scale=0.9]{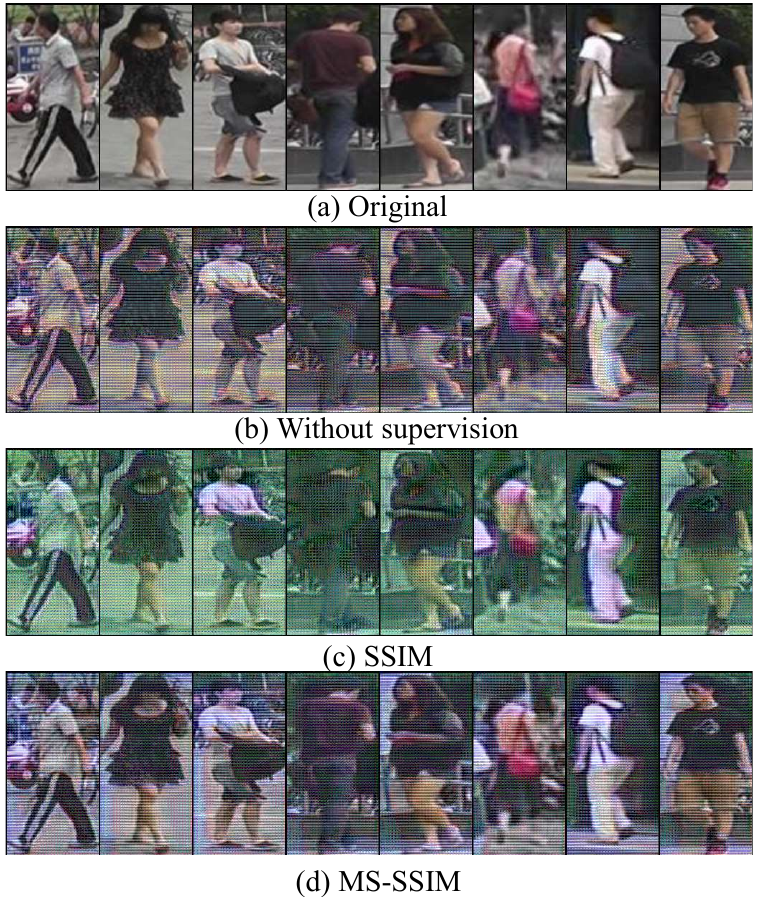}
\caption{\small{Visual comparison of using different supervisions.}}\label{fig:msssim}
\vspace{-16pt}
\end{figure}

\textbf{Using MS-SSIM.} To demonstrate the superiority of MS-SSIM, we visualize the adversarial examples under different perception supervisions in Fig. \ref{fig:msssim}. We can see that at the same magnitude, the adversarial example generated under the supervision of MS-SSIM are much better than those generated under the supervision of SSIM and without any supervision. This comparison verifies that MS-SSIM is critical to reserve the raw appearance.

\begin{table}
\centering
\caption{\small{Effectiveness of our multi-shot sampling.}}\label{tab:noise}
\scriptsize
\begin{tabular}{c|cc|ccc}
& \multicolumn{2}{c|}{\textbf{(A) random location}} & \multicolumn{2}{c}{\textbf{(B) our learned location}} \\\hline
        & R-1 & mAP & R-1 & mAP \\ \hline
Gaussian noise & 81.9 & 68.1 & 79.4 & 65.3 \\
Uniform noise & 51.1 & 40.1 & 50.7 & 39.2 \\
\textbf{Ours} & - & - & \textbf{39.7} & \textbf{30.7}
\end{tabular}
\vspace{-22pt}
\end{table}

\textbf{Comparisons of Different $\epsilon$.} Although using perception loss has great improvement for visual quality with large $\epsilon$, we also provide baseline models with smaller $\epsilon$ for full studies. We manually control $\epsilon$ by considering it as a hyperparameter. The comparisons of different $\epsilon$ are reported in Table \ref{tab:ablation} (b). Our method has achieved good results, even when $\epsilon=15$. The visualization of several adversarial examples with different $\epsilon$ can be seen in Appendix \textcolor{red}{D}.

\textbf{Number of the Pixels to be Attacked.} Let $H$ and $W$ denote the height and the width of the image. We control the number of the pixels to be attacked in the range of \{1, 1/2, 1/4, 1/8, 1/16, 1/32, 1/64\} $\times HW$ respectively by using Eqn. \ref{eqw:mask}. We have two major observations from Table \ref{tab:num}. \textbf{First}, the attack is definitely successful when the number of the pixels to be attacked $>\frac{HW}{2}$. This indicates that we can fully attack the ReID system by using a noise number of only $\frac{HW}{2}$. \textbf{Second}, when the number of pixels to be attacked $<\frac{HW}{2}$, the success attack rate drops significantly. To compensate for the decrease in noise number, we propose to enhance the noise magnitude without significantly affecting the perception. In this way, the least number of pixels to be attacked is reduced to $\frac{HW}{32}$, indicating that the number and the magnitude of the noise are both important.

\textbf{Effectiveness of Our Multi-shot Sampling.} To justify the effectiveness of our learned noise in attacking ReID, we compare them with random noise under the restriction of $\varepsilon=40$ in two aspects in Table \ref{tab:noise}. \textbf{(A)} Random noise is imposed on random locations of an image. The results suggest that rand noise is inferior to our learned noise. \textbf{(B)} Random noise is imposed on our learned location of an image. Interestingly, although (B) has worse attack performance than our learned noise, (B) outperforms (A). This indicates our method successfully finds the sensitive location to attack.

\begin{figure}
\centering
\includegraphics[scale=1.0]{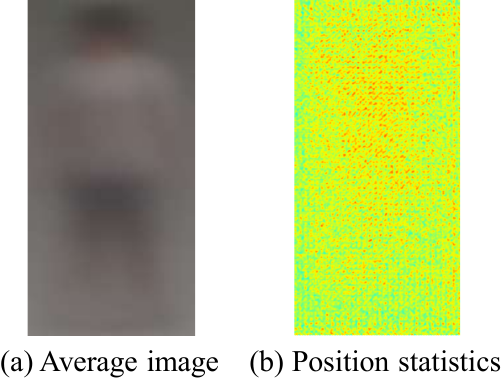}
\vspace{-4pt}
\caption{\small{\textbf{Left}: The average image of all queries on Market1501. \textbf{Right}: The frequency of adversarial points appears at different positions among Market1501 when ratio=1/8. The higher the color temperature is, the frequently the position tends to be selected.}}\label{fig:position2}
\vspace{-7pt}
\end{figure}

\emph{\textbf{Interpretability of Our Attack.} After the analysis of the superiority of our learned noise, we further visualize the noise layout to explore the interpretability of our attack in ReID. Unfortunately, a single image cannot provide intuitive information (see Appendix \textcolor{red}{B}). We statistically display query images and masks when noise number equals to $\frac{HW}{8}$ in Fig. \ref{fig:position2} for further analysis. We can observe from Fig. \ref{fig:position2} (b) that the network has a tendency to attack the top half of the average image, which corresponds to the upper body of a person in Fig. \ref{fig:position2}(a). This implies that the network is able to sketch out the dominant region of the image for ReID. For future improvement of the robustness of ReID systems, attention should be paid to this dominant region.}

\subsection{Black-Box Attack} \label{sect:blackbox}
Different from the above white-box attack, a black-box attack denotes that the attacker has no access to the training data and target models, which is very challenging.

\textbf{Cross-dataset attack.} Cross-dataset denotes that the attacker is learned on a known dataset, but is reused to attack a model that is trained on an unknown dataset. Table \ref{tab:ablation} (d) shows the success of our cross-dataset attack in AlignedReID \cite{DBLP:journals/corr/abs-1711-08184}. We also observe that the success rate of the cross-dataset attack is almost as good as the naive white-box attack. Moreover, MSMT17 is a dataset that simulates the real scenarios by covering multi-scene and multi-time. Therefore, the successful attack on MSMT17 proves that our method is able to attack ReID systems in the real scene without knowing the information of real-scene data.

\textbf{Cross-model attack.} Cross-model attack denotes that the attacker is learned by attacking a known model, but is reused to attack an unknown model. Experiments on Market1501 show that existing ReID systems are also fooled by our cross-model attacked (Table \ref{tab:ablation} (e)). It is worth to mention that PCB seems to be more robust than others, indicating that hiding the testing protocol benefits the robustness.

\textbf{Cross-dataset-cross-model attack.} We further examine the most challenging setting, i.e., our attacker has no access to both the training data and the model. The datasets and models are randomly chosen in Table \ref{tab:ablation} (f). Surprisingly, we have observed that our method has successfully fooled all the ReID systems, even in such an extreme condition. Note that Mudeep has been attacked by only 4,000 pixels.

\emph{\textbf{Discussion.} We have the following remarks for future improvement in ReID. {\textbf{First}}, although the bias of data distributions in different ReID datasets reduces the accuracy of a ReID system, it is not the cause of security vulnerability, as is proved by the success of cross-dataset attack above. {\textbf{Second}}, the success of cross-model attack implies that the flaws of networks should be the cause of security vulnerability. {\textbf{Third}}, the success of a cross-dataset-cross-model attack drives us to rethink the vulnerability of existing ReID systems. Even we have no prior knowledge of a target system; we can use the public available ReID model and datasets to learn an attacker, using which we can perform the cross-dataset-cross-model attack in the target systems. Actually, we have fooled a real-world system (see Appendix \textcolor{red}{D}).}

\begin{table}[!ht]
\centering
\caption{\small{Accuracy after non-targeted white-box attacks on CIFAR10. \emph{Original:} the accuracy on clean images. \emph{DeepFool:} \cite{moosavi2016deepfool}; \emph{NewtonFool:} \cite{jang2017objective}; \emph{NewtonFool:} \cite{jang2017objective}; \emph{CW:} \cite{carlini2017towards}; \emph{GAP:} \cite{DBLP:conf/ijcai/XiaoLZHLS18}; }}\label{tab:clf_comp}
\scriptsize
\begin{tabular}{l|c|c|c|c}
Method & \multicolumn{4}{c}{Accuracy (\%)} \\ \hlineB{3.5}
Original & \multicolumn{4}{c}{90.55} \\ \hlineB{2.5}
DeepFool& \multirow{5}{*}{$\varepsilon=8$} &58.22 & \multirow{5}{*}{$\varepsilon=2$} & 58.59 \\
NewtonFool& &69.79 & & 69.32 \\
CW & &52.27 && 53.44\\
GAP & &51.26 && 51.8\\ \clineB{1-1}{2.5}\clineB{3-3}{2.5} \clineB{5-5}{2.5}
\textbf{Ours} & & \textbf{47.31} && 50.3
\end{tabular}
\vspace{-11pt}
\end{table}

\subsection{Comparison with Existing Attackers}
To show the generalization capability of our method, we perform an additional experiment on image classification using CIFAR10. We compare our method with four advanced white-box attack methods in adversarial examples community, including DeepFool \cite{moosavi2016deepfool}, NewtonFool \cite{jang2017objective}, CW \cite{carlini2017towards}, and GAP \cite{DBLP:conf/ijcai/XiaoLZHLS18}. We employ adversarially trained ResNet32 as our target model and fix $\varepsilon=8$. Other hyper-parameters are configured using default settings the same as \cite{art2018}. For each attack method, we list the accuracy of the resulting network on the full CIFAR10 \emph{val} set. The results in Table \ref{tab:clf_comp} imply that our proposed algorithm is also effective in obfuscating the classification system. Note that changing $\varepsilon$ to other numbers (e.g., $\varepsilon=2$) does not reduce the superiority of our method over the competitors.%

\section{Conclusion}
We examine the insecurity of current ReID systems by proposing a learning-to-mis-rank formulation to perturb the ranking of the system output. Our mis-ranking based attacker is complementary to the existing misclassification based attackers. We also develop a multi-stage network architecture to extract general and transferable features for the adversarial perturbations, allowing our attacker to perform a black-box attack. We focus on the inconspicuousness of our attacker by controlling the number of attacked pixels and keeping the visual quality. 
The experiments not only show the effectiveness of our method but also provides directions for the future improvement in the robustness of ReID.

\section{Appendix A: More Visualization of the Destruction to the ReID system}
\begin{figure}[!htb]
\centering
\includegraphics[width=1 \columnwidth]{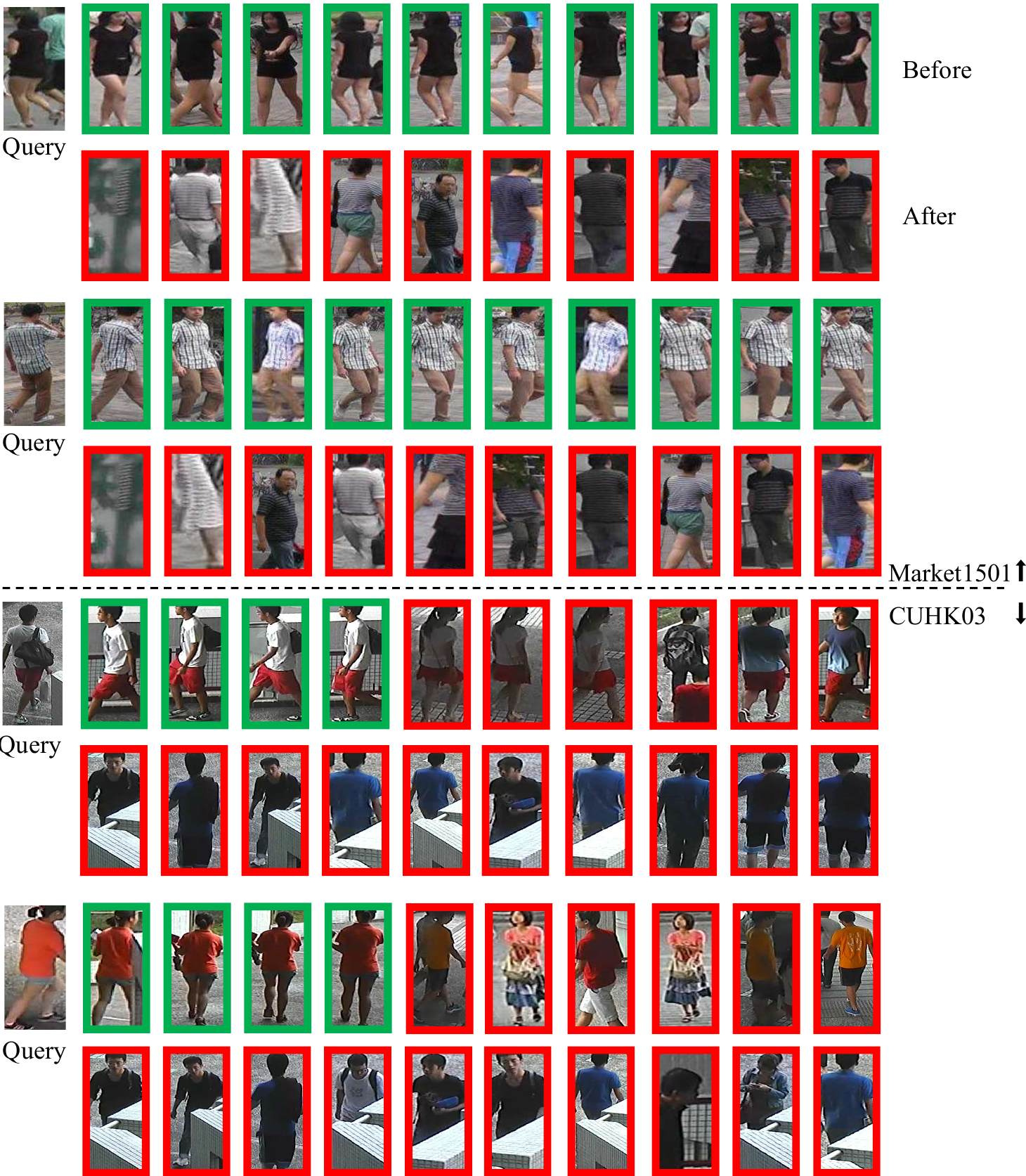}
\caption{The rank-10 predictions of AlignedReID\cite{sun2018beyond} (one of the state-of-the-art ReID models) before and after our attack on Market-1501 and CUHK03. We display the gallery images according to their rank accuracies returned from AlignedReID\cite{DBLP:journals/corr/abs-1711-08184} model. The green boxes are the correctly matched images, while the red boxed are the mismatched images. Only top-10 gallery images are visualized.}
\label{rank10}
\end{figure}

We present more results from both Market1501 and CUHK03 datasets by exhibiting the Rank-10 matches from the target ReID system before and after an adversarial attack in Figure \ref{rank10}.

\section{Appendix B: Visualization of Sampling Mask}
\begin{figure}[!htb]
\centering
\includegraphics[width=1 \columnwidth]{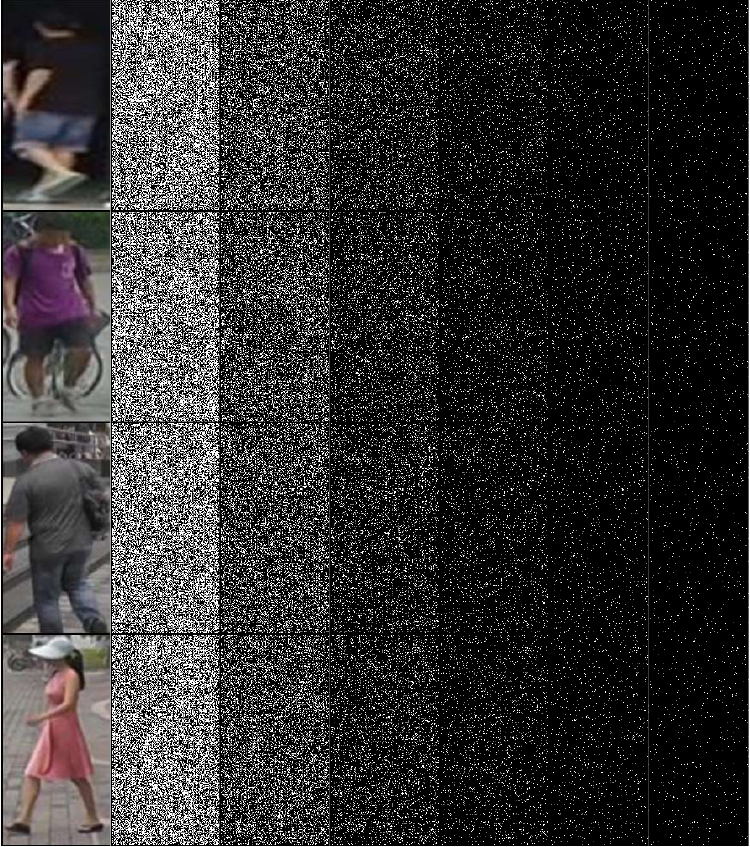}
\caption{The layout of the noise. Column 1 shows the query images. Column 2-7 are the noise layouts with different noise numbers, i.e., \{1/2, 1/4, 1/8, 1/16, 1/32, 1/64\} $\times HW$.}
\vspace{-11pt}
\label{fig:position1}
\end{figure}

We further visualize the noise layout in Figure \ref{fig:position1} to explore the interpretability of our attack in ReID. However, it is hard to get intuitive information from a few sporadic samples.

\section{Appendix C: More Details of Attacking}
The experiments are performed on a workstation with an NVIDIA Titan X GPU with 12GB GPU memory. The protocols of target models are the same as their official releases. The minibatch size is $32$ for 50 epochs. The filters in each multi-column network in our multi-stage discriminator are $4\times 4$. The weights were initialized from a truncated normal distribution with mean $0$ and standard deviation $0.02$ if the layer is not followed by the spectral normalization. The whole network is trained in an end-to-end fashion using Adam \cite{DBLP:journals/corr/KingmaB14} optimizer with the default setting, and the learning rate is set to 2e-4.

As for the balanced factors in our full objective, the selection of $\beta$ mainly depends on how worse the visual quality we can endure. In our case, we fix $\alpha=2$ and $\beta=5$ in the main experiments but $\beta$ ranging from $5$ to $0$ when we gradually cut down the number of attacked pixels.

\section{Appendix D: Attacking to the Real-world System}
\begin{figure}[!htb]
\centering
\includegraphics[width=1\columnwidth]{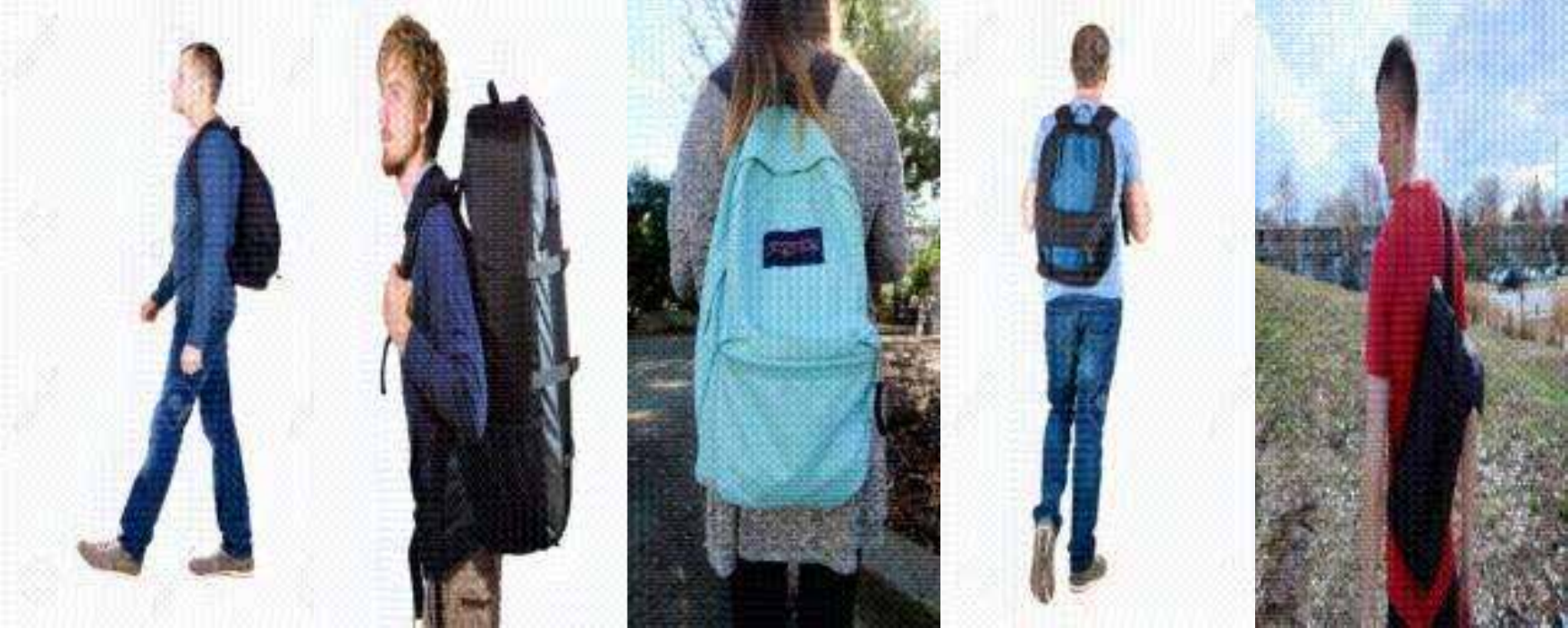}
\caption{Visualization of several adversarial examples with $\varepsilon$=20,20,30,30,40,40 respectively which successfully attack Baidu AI service.}
\label{fig:atk_baidu}
\end{figure}

To the best of our knowledge, no ReID API is available online for the public to attack. Fortunately, we find a real-world system, \emph{\emph{Human Detection and Attributes} API provided by Baidu (https://ai.baidu.com/tech/body/attr)}, that can examine the capacity of our attacker. Actually, our attacker is able to perform an attribute attack, which is crucial in ReID. We randomly pick up person images \emph{with backpack} from Google and generate noise to them using our attacker. This results in the adversarial examples listed in Fig. \ref{fig:atk_baidu}. When uploading these adversarial examples to the Baidu API, the API misclassified them as persons \emph{without backpack}, implying the real-world system has been fooled by our attacker.

\section*{Acknowledgement}
\noindent
This work was supported in part by the State Key Development Program (No. 2018YFC0830103), in part by NSFC (No.61876224,U1811463,61622214,61836012,61906049), and by GD-NSF (No.2017A030312006,2020A1515010423).

{\small
\bibliographystyle{ieee_fullname}
\bibliography{egbib}
}

\end{document}